# Exploring Traffic Crash Narratives in Jordan Using Text Mining Analytics

Shadi Jaradat
*CARRS-Q*
*Queensland University of Technology*
Brisbane, Australia
shadi.jaradat@hdr.qut.edu.au

Taqwa I. Alhadidi
*Civil Engineering Department*
*Al-Ahliyya Amman University*
Amman, Jordan
t.alhadidi@ammanu.edu.jo

Huthaifa I. Ashqar
*Civil Engineering Department*
*Arab American University*
Jenin, Palestine
huthaifa.ashqar@aaup.edu, 0000-0002-6835-8338

Ahmed Hossain
*Department of Civil Engineering*
*University of Louisiana at Lafayette*
Lafayette, Louisiana
ahmed.hossain1@louisiana.edu

Mohammed Elhenawy
*CARRS-Q*
*Queensland University of Technology*
Brisbane, Australia
mohammed.elhenawy@qut.edu.au

*Abstract*—This study explores traffic crash narratives in an attempt to inform and enhance effective traffic safety policies using text-mining analytics. Text mining techniques are employed to unravel key themes and trends within the narratives, aiming to provide a deeper understanding of the factors contributing to traffic crashes. This study collected crash data from five major freeways in Jordan that cover narratives of 7,587 records from 2018-2022. An unsupervised learning method was adopted to learn the pattern from crash data. Various text mining techniques, such as topic modeling, keyword extraction, and Word Co-Occurrence Network, were also used to reveal the co-occurrence of crash patterns. Results show that text mining analytics is a promising method and underscore the multifactorial nature of traffic crashes, including intertwining human decisions and vehicular conditions. The recurrent themes across all analyses highlight the need for a balanced approach to road safety, merging both proactive and reactive measures. Emphasis on driver education and awareness around animal-related incidents is paramount.

## I. Introduction

Recent advancements in text analysis and natural language processing (NLP) have opened new avenues for extracting valuable insights from crash narratives [1]. These narratives provide thorough accounts of pre-crash movements, driver behaviors, road conditions, and other contextual aspects frequently missing from traditional traffic crash reports. Researchers can find underlying patterns and contributing reasons to crashes by examining these events, which may not be discernable through quantitative data alone [1], [2], [3].

The use of crash narratives in traffic crash modeling is a significant step toward a more comprehensive technique of traffic safety analysis. This method takes into account the complexities of environmental interactions, situational variables, and human factors that contribute to traffic crashes. Narrative analysis, for example, can uncover recurring themes in crashes, such as driver distraction, road anger, or specific environmental hazards, which can then be used to design targeted solutions [4], [5], [6].

This This study employs text mining analytics to delve into traffic crash narratives with the aim of informing and improving traffic safety policies. Using data from five major freeways in Jordan spanning 2018-2022, the study analyzes 7,587 crash records. Unsupervised learning methods uncover patterns, while text mining techniques such as topic modeling, keyword extraction, and Word Co-Occurrence Network reveal co-occurring crash patterns.

## II. Literature Review

Previous studies provided valuable insights into crash story modeling using Natural Language Processing (NLP) and sophisticated language models. Researchers apply advanced language models to categorize pedestrian crash types, demonstrating NLP's promise in crash typing tools. They highlight narration inconsistencies, data imbalance, and small sample sizes as potential obstacles to growth in this field [7], [8]. Another study also uses negative sampling in spatio-temporal crash prediction, highlighting NLP's role in network-level crash analysis. This study shows the wider influence of NLP on crash prediction and analysis beyond crash narrative modeling [9]. Chen and Tao (2022) used text mining to examine expressway traffic crash duration and discovered that it can help explain traffic crash causes [10] Other work described a novel way of assessing road-curve collision severity using text mining, highlighting the potential of text mining to improve risk and crash severity information [11].

Various methods analyzed highway safety and crash severity. XGBoost and Bayesian networks have been used to forecast highway crash severity. These models consider road, environmental, real-time traffic, and meteorological issues. The XGBoost model determines crash severity-affecting factors [12], [13], [14]. Select variables and their values are used to build the Bayesian network-based model to predict crash severity [13] These models perform well in learning and prediction. The interaction effect of road and environment elements is also studied, improving prediction performance. These models can theoretically support motorway safety management and improvement efforts. Himes et al. (2022) developed a safety prediction approach for high-occupancy lanes on motorways using models for total and multiple-vehicle crash frequency. This thorough approach to analysing safety in high-occupancy lanes on motorways provides insights into crash frequency and severity [15].

Safety model transferability to other motorway networks has also been investigated. Examined the transferability of the Highway Safety Manual Motorway Model to the Italian Motorway Network, focusing on crash severity-based calibration parameters for motorway segments and speed-change lanes [16], [17]. This implies that safety models should take into account motorway section characteristics . Various elements are considered while assessing motorway safety.

suggested a holistic motorway traffic safety rating methodology that considers subjective and objective elements [18]. Additionally, machine learning to analyze and predict crash injury severity based on contributing factors, showing that advanced analytical tools may help us comprehend crash severity causes [19], [20] A study found that animals and low sight from lampless roads enhance traffic crash risk. Male drivers often crash in the early hours (12–6 am) without streetlight or passengers. Male drivers cause drug- and alcohol-related roadside crashes. Alcohol and drugs are likely to affect highway crash drivers [21].

## III. METHODOLOGY

### A. Data Collection

This study examines the research question: "What patterns can be identified from traffic crash narratives to inform and enhance effective traffic safety policies?" In pursuit of this goal, the research delves into analyzing crash narratives through various text-mining techniques. These methods are employed to unravel key themes and trends within the narratives, aiming to provide a deeper understanding of the factors contributing to traffic crashes. This study collected data from five major freeways in Jordan, namely Airport Road, Desert Highway, Jordan Highway Route 30, and Route 35. The collected crash data covers a narrative of 7,587 crash records from 2018-2022. Before starting the text mining techniques, several essential data-cleaning steps were performed. This involves checking for and addressing any missing values, cleaning the data by removing stop words, special characters, and irrelevant symbols, tokenizing the text into words, and converting all text to lowercase for consistency. An unsupervised learning method was adopted to learn the pattern from crash data. Various text mining techniques were used to reveal the co-occurrence of crash patterns, such as topic modeling, keyword extraction, and Word Co-occurrence Network. Fig. 1 shows a brief overview of the proposed methodology.

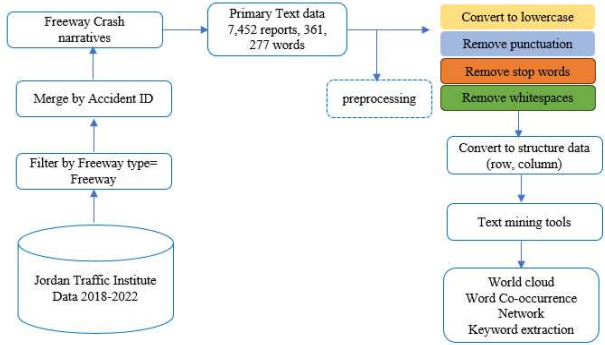

Fig. 1. Flowchart of the proposed method.

### B. Topic Modeling

Latent Dirichlet allocation (LDA) is an unsupervised topic model, a widely used probabilistic clustering algorithm to determine thematic information within extensive text datasets. This methodology is used to reduce the dimensionality of the corpus using its latent correlations [22], [23]. This modeling approach is structured around the concept of probability distributions related to topics and the words associated with them. The basic concept involves representing documents as a stochastic blend of underlying topics, with each topic being a probability distribution across words.

### C. Word Co-occurrence Network (WCN)

Another approach to comprehending patterns within unstructured crash narratives involves examining the co-occurrence of terms. This research conducted co-occurrence analysis across the entire corpus (i.e., the crash narrative data) (48). The subsequent section highlights some noteworthy term co-occurrences as appeared in Figure 2. A Word Co-occurrence Network (WCN) is created by establishing connections between vertices that correspond to n consecutive words within a sentence, utilizing the concept of n-grams.

### D. Rapid Automatic Keyword Extraction (RAKE)

The Rapid Automatic Keyword Extraction (RAKE) algorithm is an unsupervised and language-independent technique for extracting keywords from individual documents. First, the text is segmented into a collection of candidate keywords. This collection is further divided into sequences of contiguous words at stop word locations and phrase delimiters. After identifying the candidate keywords, a score is calculated as the sum of the scores of its constituent member keywords. The highest-ranking RAKE keywords are subsequently determined based on their computed scores. Following the scoring of candidate keywords, the top-scoring candidates are chosen as keywords for the document [24]

## IV. ANSLYSIS AND RESULTS

### A. Word Cloud Analysis

In the word cloud visualization of traffic crash narratives, shown in Fig. 2, a systematic assessment reveals several focal themes that delineate the multifaceted nature of vehicular incidents. Prominent terms such as "collision," "hit," "violation," and "impact" signify the predominance of direct physical interactions, likely between vehicles or with stationary objects. Concurrently, the emergence of terms like "violation," "error," and "noncompliance" underscores the pivotal role of human-induced factors, suggesting that breaches in traffic regulations or lapses in driver judgment frequently accelerate these incidents. Mechanical aspects are underscored by terms such as "mechanical," "leaking," and specific vehicular components like "wheels" and "mirror." The presence of these terms might indicate a correlation between vehicular malfunctions or deteriorations and the propensity for crashes. Furthermore, environmental elements are subtly highlighted, with references to "water," "light," "direction," and "noon," pointing towards the influential role of external conditions in vehicular dynamics and crash risks. The recurring emphasis on "safety," juxtaposed with more dire outcomes denoted by "death" and "injury," accentuates the profound human toll and underscores the exigency for enhanced preventive measures. Collectively, this visualization offers a holistic perspective on the myriad factors contributing to traffic crashes, providing valuable insights for future research and policy formulation.

### B. Word Co-occurrence Network (WCN)

The word occurrence network provides a holistic overview, underscoring the multifactorial nature of traffic crashes. The intricate interconnections between terms offer invaluable insights, emphasizing both human-induced and environmental risk factors. Such a detailed understanding paves the way for targeted interventions and policy formulation to enhance road safety. Several observations can be made as shown in Fig. 3.

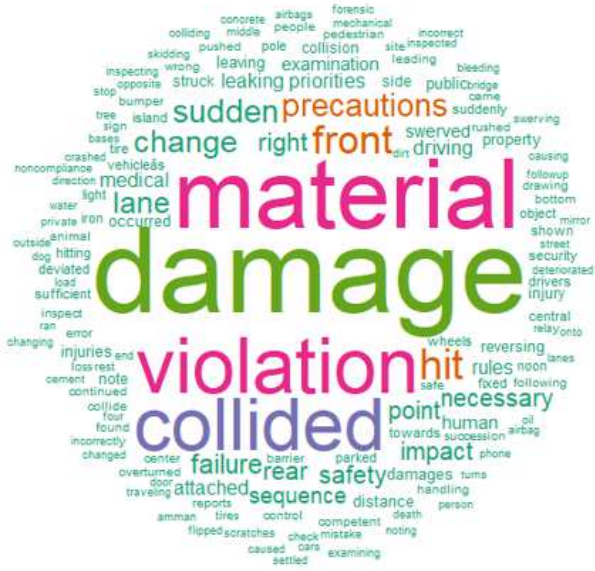

Fig. 2. World Cloud visualization

The association between "Dog" and "Animal" underscores the potential hazards posed by animals on roads. In fact, this becomes noticed as the Jordanian government issues a role of prohibiting killing stray animals. Also, this suggests that abrupt encounters with fauna might precipitate unforeseen circumstances for drivers which is agree with the results of [21]. The trifold linkage among "Sudden," "Change," and "Leaking" intimates the unexpected nature of vehicular malfunctions or conditions that could lead to crashes.

A noteworthy connection is "Leaving-Distance-Sufficient," emphasizing the critical importance of maintaining adequate spacing between vehicles to prevent collisions. This is further reinforced by the explicit pairing of "Collision" and "Occurred," signifying the central theme of direct physical interactions in the narratives. The triadic relation "Left-Rear-Front-Wheel" provides granularity on vehicular dynamics, hinting that specific parts of vehicles, especially wheels, play pivotal roles in crash causation or outcomes.

Furthermore, the association between "Opposite" and "Direction" coupled with "Barrier-Cement" suggests potential scenarios where vehicles might deviate into oncoming traffic or barriers, possibly due to obstructions or errors. The coupling of terms such as "Sign," "Stop," and "Noncompliance" underscores a recurrent theme: the significance of adhering to traffic regulations to ensure road safety. Environmental conditions are subtly emphasized, with "Water-Oil-Bleeding" indicating the role of road surfaces and conditions in crashes, while "Phone-Malfunction-Misuse" highlights human-induced risks associated with distractions or technological failures.

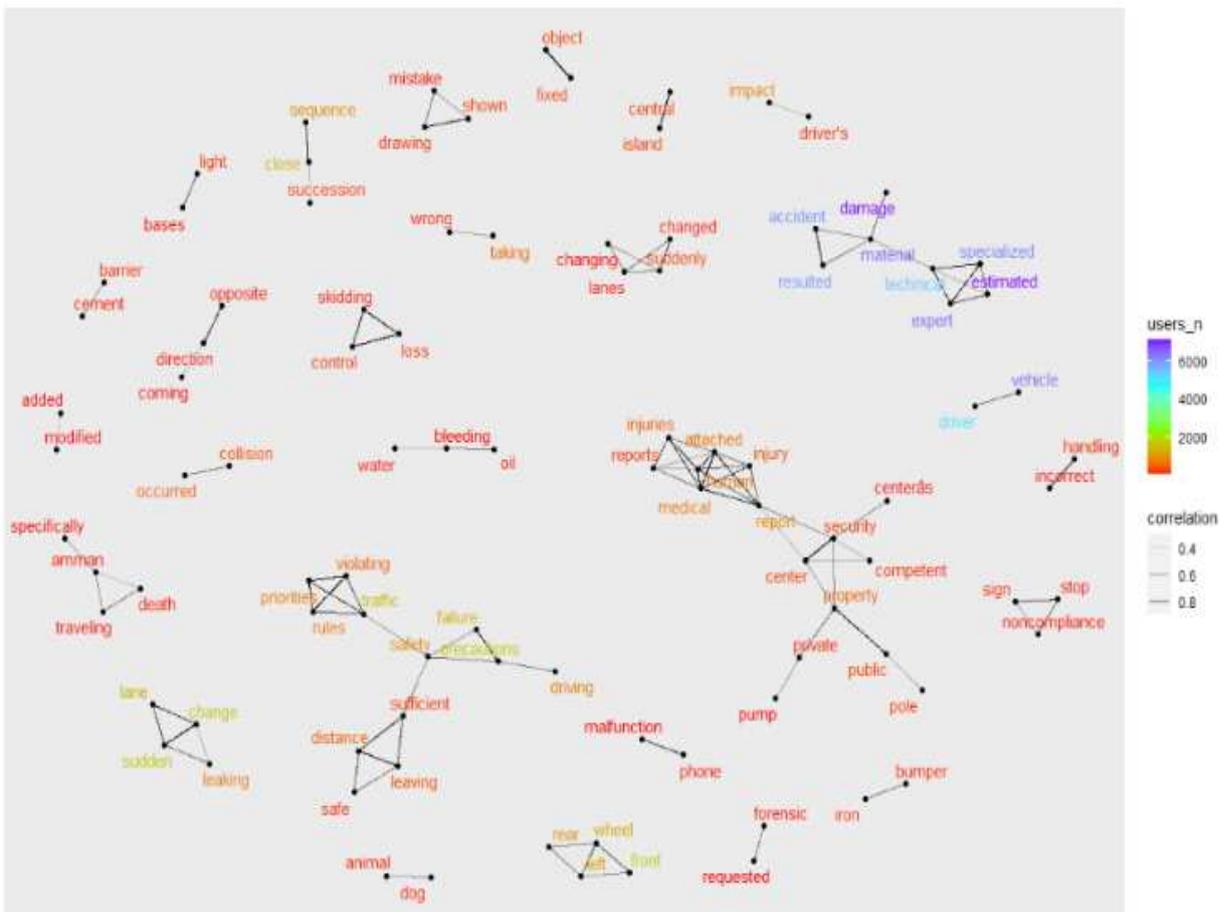

Fig. 3. Word Co-occurrence Network (WCN) Analysis

## C. Rapid Automatic Keyword Extraction (RAKE)

The RAKE analysis of traffic crash narratives unveiled a constellation of pivotal factors underpinning vehicular incidents, as shown in Fig. 4. Key terms such as "Material Damage" and "Vehicle Failure" underscore the tangible impacts and mechanical challenges often faced on the road.

Environmental elements emerged prominently, with terms like "Oil Bleeding" and "Concrete Debris" highlighting potential road hazards, while the mention of "Animal Sheep" indicates interactions with fauna as noteworthy contributors to certain incidents. Furthermore, human behaviors and decision-making processes play a central role, as evidenced by terms like "Reckless Manner" and "Traffic Rules." The extracted keywords collectively paint a multifaceted picture of the complexities surrounding traffic crashes, emphasizing the intertwined nature of human, vehicular, and environmental elements in shaping road safety dynamics. Some key findings from RAKE analysis include:

1- Vehicle Components and Damage
   a. Material Damage: Indicates tangible harm to vehicles or infrastructure.
   b. Mud Flap: Specific vehicle parts that might be involved or damaged in incidents. Tire Explosion: Points to a significant mechanical failure that can lead to crashes. Vehicle Failure: A broader term that suggests mechanical or electronic issues with a vehicle.
2- Objects and Obstructions
   a. Light Pole & Electricity Pole: Stationary objects that vehicles might collide with.
   b. Concrete Debris & Fixed Object: Indicates potential road obstructions.
   c. Waste Container: Suggests incidents might occur near areas with waste disposal or storage.
3- Driving Behavior and Violations
   a. Opposite Lane & Wrong Lane: Indicates potential issues with lane discipline or vehicles being in places they shouldn't be.
   b. Traffic Rules: A general term that underscores the importance of adhering to regulations.
   c. Violation Failure: This suggests that not just violations but failures to recognize or act on them play a role in incidents.
   d. Reckless Manner: Points to aggressive or careless driving behavior.
4- Environmental and External Factors
   a. Oil Bleeding: This can refer to vehicle leaks or spills on the road that make surfaces slippery.
   b. Animal Sheep: Indicates that animals, in this case specifically sheep, can be hazards on the road.
   c. Airport Road: Suggests that specific locations, like roads near airports, might have unique risk factors.
   d. Health Condition: This might point to incidents where the driver's health condition played a role, e.g., medical emergencies while driving.
5- Driving Maneuvers and Responses
   a. Rear End: Refers to a type of collision, typically when one vehicle hits the back of another.
   b. Sudden Change: Indicates abrupt maneuvers or changes in driving conditions.

## D. Topic Modeling

In our analysis of topic modeling for the crash narrative corpus, we utilized a coherence score to determine the optimal number of topics represented by k. Results indicated a peak coherence score of 0.1 when k equals 7. This suggests that the crash narratives can be best represented and understood by segmenting them into seven distinct topics. A coherence score of 0.1, in this context, implies a reasonable degree of semantic clarity and distinction among the seven topics. Such an optimal number ensures that the topics are neither too granular, which could result in overlapping themes, nor too broad, which might overlook nuanced differences. Thus, based on the coherence score, a seven-topic model offers the most insightful and structured breakdown of the crash narrative corpus.

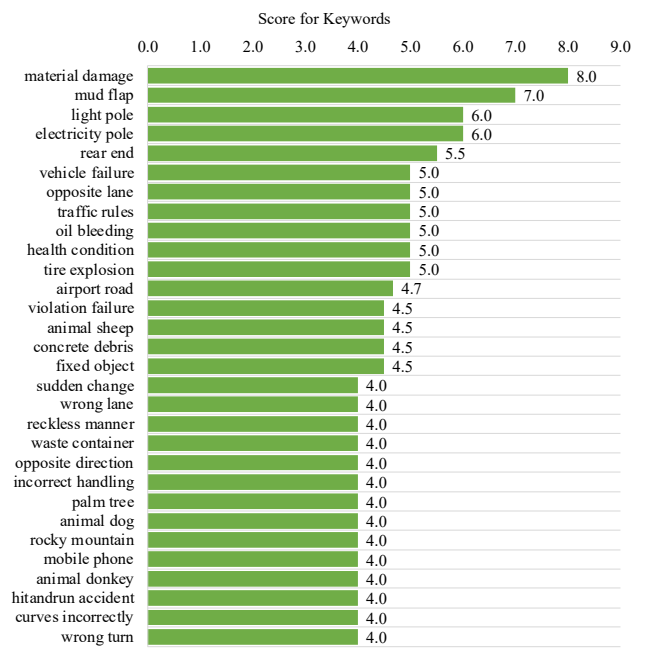

Fig. 4. RAKE Analysis

A comprehensive topic modeling analysis was undertaken on a corpus of crash narratives. This analysis distilled the vast array of narratives into seven distinct topics, each encapsulating a unique dimension of traffic crashes. These topics collectively offer a holistic lens through which the complexities of road incidents can be understood. From the technical intricacies of vehicles to the broader strokes of human behavior and traffic regulations, these topics weave a tapestry of factors that contribute to, result from, and characterize the landscape of road safety and crashes. The ensuing sections delve deep into each of these seven topics, shedding light on their significance and the insights they offer.

The topic modeling of the crash narrative corpus has elucidated seven distinct thematic areas. Emphasis on vehicle dynamics, technical evaluations, and specific damages is recurrent across multiple topics, indicating their central role in crash narratives. Notably, the analysis underscores the intertwined nature of human decisions, vehicle conditions, and external factors in shaping the dynamics of road incidents. From the meticulous assessment of damages (as seen in Topics 2 and 3) to the adherence (or lack thereof) to traffic regulations (Topic 4), the results offer a comprehensive panorama of crash causation and consequences. Furthermore, the emphasis on documentation and the human element in Topic 7 accentuates the profound personal and administrative impacts of road incidents. Collectively, these topics provide a granular and multifaceted understanding of the complexities surrounding traffic crashes.

**Topic 1. Safety & Consequences:** This topic seems to focus on the preventive measures and consequences of crashes. Words like "precautions," "failure," and "result" suggest the importance of safety measures and the potential outcomes if they aren't heeded. The presence of "expert" and "material" might indicate evaluations or assessments post-incident.

**Topic 2. Crash Assessment:** Emphasizing on the aftermath of a crash, this topic brings in words like "estimated," "expert," and "violation," hinting at the evaluation of damages, both in terms of costs and potential traffic rule breaches.

**Topic 3. Vehicle Dynamics & Damage:** Highlighting specific parts of vehicles, such as "front," "wheel," "left," "rear," and "right," this topic likely revolves around the dynamics of crashes and how different parts of vehicles are impacted.

**Topic 4. Traffic Regulations & Priorities:** With words like "traffic," "rules," and "priorities," this topic underscores the role of traffic regulations and the importance of adhering to them to prevent crashes.

**Topic 5. Technical Aspects & Collision:** This topic, with terms like "technical," "driver," and "collided," seems to revolve around the technical aspects of vehicles, driver behavior, and the nature of collisions.

**Topic 6. Change & Unforeseen Events:** The presence of "change" and "sudden" in this topic hints at unexpected events or maneuvers leading to crashes, emphasizing the unpredictable nature of some incidents.

**Topic 7. Documentation & Human Element:** This topic, with words like "report," "medical," "attached," and "human," seems to focus on the documentation post-crash and the human aspect, be it injuries or the human factors contributing to the crash.

## V. CONCLUSION

The multifaceted analysis comprising word cloud visualization, RAKE, and topic modeling has unveiled a rich tapestry of themes and associations related to traffic crashes. The word cloud analysis spotlighted immediate terms, indicating the tangible repercussions of crashes, such as "material," "damage," and "violation." This was further nuanced by the RAKE results, which brought specific scenarios and intricacies, like "Vehicle Failure" and "Reckless Manner."

Topic modeling provided a structured breakdown of the narratives, categorizing them into seven distinct thematic areas. These topics ranged from the technical aspects of vehicles to human behaviors and emphasized the reactive nature of our post-crash management systems.

Collectively, these analyses underscore the multifactorial nature of traffic crashes, intertwining human decisions, vehicular conditions, environmental elements, and more. The recurrent themes across all analyses highlight the need for a balanced approach to road safety, merging both proactive and reactive measures. Emphasis on driver education, enhanced infrastructure, regular vehicle maintenance, and awareness around animal-related incidents is paramount. Moreover, the insights suggest that leveraging modern technologies for real-time monitoring, reporting, and predictive analytics can play a transformative role in minimizing risks.

REFERENCES

[1] S. Das, A. H. Oliaee, M. Le, M. P. Pratt, and D. Wu, "Classifying Pedestrian Maneuver Types Using the Advanced Language Model," *Transportation Research Record Journal of the Transportation Research Board*, 2023, doi: 10.1177/03611981231155187.

[2] R. L. Rose, T. G. Puranik, and D. N. Mavris, "Natural Language Processing Based Method for Clustering and Analysis of Aviation Safety Narratives," *Aerospace*, 2020, doi: 10.3390/aerospace7100143.

[3] Y. Weng, S. Das, and S. G. Paal, "Applying Few-Shot Learning in Classifying Pedestrian Crash Typing," *Transportation Research Record Journal of the Transportation Research Board*, 2023, doi: 10.1177/03611981231157393.

[4] A. J.-P. Tixier, M. R. Hallowell, B. Rajagopalan, and D. Bowman, "Automated content analysis for construction safety: A natural language processing system to extract precursors and outcomes from unstructured injury reports," *Autom Constr*, vol. 62, pp. 45–56, 2016.

[5] N. O. Khanfar, M. Elhenawy, H. I. Ashqar, Q. Hussain, and W. K. M. Alhajyaseen, "Driving behavior classification at signalized intersections using vehicle kinematics: Application of unsupervised machine learning," *Int J Inj Contr Saf Promot*, pp. 1–11, 2022.

[6] N. O. Khanfar, H. I. Ashqar, M. Elhenawy, Q. Hussain, A. Hasasneh, and W. K. M. Alhajyaseen, "Application of Unsupervised Machine Learning Classification for the Analysis of Driver Behavior in Work Zones in the State of Qatar," *Sustainability*, vol. 14, no. 22, p. 15184, 2022.

[7] S. Das, A. H. Oliaee, M. Le, M. P. Pratt, and D. Wu, "Classifying Pedestrian Maneuver Types Using the Advanced Language Model," *Transportation Research Record Journal of the Transportation Research Board*, 2023, doi: 10.1177/03611981231155187.

[8] Y. Weng, S. Das, and S. G. Paal, "Applying Few-Shot Learning in Classifying Pedestrian Crash Typing," *Transportation Research Record: Journal of the Transportation Research Board*, p. 036119812311573, Mar. 2023, doi: 10.1177/03611981231157393.

[9] P. Way, J. Roland, M. Sartipi, and O. A. Osman, "Spatio-Temporal Crash Prediction: Effects of Negative Sampling on Understanding Network-Level Crash Occurrence," *Transportation Research Record Journal of the Transportation Research Board*, 2021, doi: 10.1177/0361198121991836.

[10] J. Chen and W. Tao, "Traffic Accident Duration Prediction Using Text Mining and Ensemble Learning on Expressways," *Sci Rep*, 2022, doi: 10.1038/s41598-022-25988-4.

[11] A. Rakotonirainy, S. Chen, B. Scott-Parker, S. W. Loke, and S. Krishnaswamy, "A Novel Approach to Assessing Road-Curve Crash Severity," *Journal of Transportation Safety & Security*, 2014, doi: 10.1080/19439962.2014.959585.

[12] T. Hasan, M. Abdel-Aty, and N. Mahmoud, "Freeway crash prediction models with variable speed limit/variable advisory speed," *J Transp Eng A Syst*, vol. 149, no. 3, p. 04022159, 2023.

[13] C. Li, X. Wu, Z. Zhang, Z. Ma, Y. Zhu, and Y. Chen, "Freeway traffic accident severity prediction based on multi-dimensional and multi-layer Bayesian network," in *2022 IEEE 2nd International Conference on Power, Electronics and Computer Applications (ICPECA)*, IEEE, 2022, pp. 1032–1035.

[14] S. Jaradat, R. Nayak, and M. Elhenawy, "Explainable Language Models For The Identification Of Factors Influencing Crash Severity Levels In Imbalanced Datasets," in *Proceedings of the 2024 3rd International Conference on Computing and Machine Intelligence (ICMI)*, Michigan: IEEE, Apr. 2024.

[15] S. Himes, J. A. Bonneson, V. V Gayah, and C. Liu, "Safety Prediction Method for Freeway Facilities With High-Occupancy Lanes," *Transportation Research Record Journal of the* Transportation Research Board, 2022, doi: 10.1177/03611981221083918.

[16]   F. L. Torre, L. Domenichini, F. Corsi, and F. Fanfani, "Transferability of the Highway Safety Manual Freeway Model to the Italian Motorway Network," Transportation Research Record Journal of the Transportation Research Board, 2014, doi: 10.3141/2435-08.
[17]   M. Elhenawy, H. A. Rakha, and H. I. Ashqar, "Joint impact of rain and incidents on traffic stream speeds," J Adv Transp, vol. 2021, 2021.
[18]   X. Liu, "Comprehensive Evaluation Model of Freeway Traffic Safety Considering Subjective and Objective Factors," IOP Conf Ser Earth Environ Sci, 2019, doi: 10.1088/1755-1315/371/3/032036.
[19]   M. Cai, F. Tang, and X. Fu, "A Bayesian Bivariate Random Parameters and Spatial-Temporal Negative Binomial Lindley Model for Jointly Modeling Crash Frequency by Severity: Investigation for Chinese Freeway Tunnel Safety," Ieee Access, 2022, doi: 10.1109/access.2022.3165065.
[20]   H. I. Ashqar, Q. H. Q. Shaheen, S. A. Ashur, and H. A. Rakha, "Impact of risk factors on work zone crashes using logistic models and Random Forest," in *2021 IEEE International Intelligent Transportation Systems Conference (ITSC)*, IEEE, 2021, pp. 1815–1820.
[21]   S. Das, A. Dutta, and X. Sun, "Patterns of rainy weather crashes: Applying rules mining," Journal of Transportation Safety & Security, vol. 12, no. 9, pp. 1083–1105, 2020.
[22]   D. M. Blei, A. Y. Ng, and M. I. Jordan, "Latent dirichlet allocation," Journal of machine Learning research, vol. 3, no. Jan, pp. 993–1022, 2003.
[23]   M. Pavlinek and V. Podgorelec, "Text classification method based on self-training and LDA topic models," Expert Syst Appl, vol. 80, pp. 83–93, 2017.
[24]   S. Rose, D. Engel, N. Cramer, and W. Cowley, "Automatic Keyword Extraction from Individual Documents," in Text Mining: Applications and Theory, John Wiley and Sons, 2010, pp. 1–20. doi: 10.1002/9780470689646.ch1.

.